\documentclass[conference]{IEEEtran}
\usepackage{times}

\usepackage[numbers]{natbib}
\usepackage{multicol}
\usepackage[bookmarks=true]{hyperref}

\usepackage{booktabs} 
\usepackage{caption}  
\usepackage{amsmath}
\usepackage{amsfonts}

\usepackage{dsfont}
\usepackage{amssymb}
\usepackage{bm}
\usepackage{cite}
\usepackage{mathtools}
\usepackage{graphicx}
\usepackage{textcomp}
\usepackage{xcolor}
\usepackage{subfigure}
\usepackage{multirow}
\usepackage{colortbl}
\usepackage{float}
\usepackage{array}
\usepackage{newtxmath}
\usepackage{makecell} 
\usepackage{booktabs} 
\usepackage{arydshln}

\begin{document}

\title{FocusNav: Spatial Selective Attention with Waypoint Guidance for Humanoid Local Navigation}


\author{\authorblockN{Yang Zhang\textsuperscript{1,2},
Jianming Ma\textsuperscript{1,2},
Liyun Yan\textsuperscript{1,2},
Zhanxiang Cao\textsuperscript{1,2},
Yazhou Zhang\textsuperscript{1},
Haoyang Li\textsuperscript{1,2},
Yue Gao\textsuperscript{1,2}}
\authorblockA{
\textsuperscript{1}Shanghai Jiao Tong University \quad \textsuperscript{2}Shanghai Innovation Institute
}
}



%

\maketitle

\begin{abstract}
Robust local navigation in unstructured and dynamic environments remains a significant challenge for humanoid robots, requiring a delicate balance between long-range navigation targets and immediate motion stability.
In this paper, we propose FocusNav, a spatial selective attention framework that adaptively modulates the robot's perceptual field based on navigational intent and real-time stability.
FocusNav features a Waypoint-Guided Spatial Cross-Attention (WGSCA) mechanism that anchors environmental feature aggregation to a sequence of predicted collision-free waypoints, ensuring task-relevant perception along the planned trajectory.
To enhance robustness in complex terrains, the Stability-Aware Selective Gating (SASG) module autonomously truncates distal information when detecting instability, compelling the policy to prioritize immediate foothold safety.
Extensive experiments on the Unitree G1 humanoid robot demonstrate that FocusNav significantly improves navigation success rates in challenging scenarios, outperforming baselines in both collision avoidance and motion stability, achieving robust navigation in dynamic and complex environments.
\end{abstract}

\IEEEpeerreviewmaketitle

\section{Introduction}\label{Introduction}
Humanoid robots, characterized by human-like bipedal locomotion systems and dexterous whole-body coordination~\citep{gu2025humanoid, weng2025hdmi}, possess the unique potential to unify locomotion and manipulation~\citep{xue2025leverb}.
This potential enables them to exhibit human-like dynamic navigation and physical interaction within complex environments~\citep{jiang2025wholebodyvla, wei2025ground}.
However, their practical deployment remains strictly constrained by their mobile navigation capabilities~\citep{peng2025logoplanner}, specifically the ability to traverse complex spaces adaptively and safely~\citep{ben2025gallant}.
While existing methodologies have demonstrated reliable obstacle avoidance in controlled, flat environments with sparse obstacles~\citep{cai2025navdp}, their navigational robustness remains fragile when transitioning to unstructured, rugged, and densely populated real-world settings~\citep{he2025attention}.
In such sophisticated scenarios, robots must simultaneously manage dynamic obstacle avoidance and stable locomotion over uneven terrain, a dual challenge that current frameworks struggle to address effectively~\citep{lee2024learning}.
Consequently, enhancing robust mobile navigation for complex, dynamic, and unstructured environments is a core challenge and a critical bottleneck to fully unleashing the potential of humanoid robot applications~\citep{lin2025let}.

To enhance the navigational capabilities of humanoid robots, researchers have explored various methodologies, including Large Language Model (LLM)-assisted navigation~\citep{zhou2024navgpt} and end-to-end Vision-Language Navigation (VLN)~\citep{cheng2024navila}.
LLM-assisted navigation leverages exceptional semantic understanding and logical reasoning to interpret task instructions and generate high-level plans, proving particularly effective for decision-making in large-scale, mapless environments~\citep{liu2025embodied}.
Meanwhile, VLN aligns semantic and scene information through cross-modal models to achieve a direct mapping from natural language to real-time navigational actions, effectively mitigating the system latency and cumulative errors inherent in traditional modular approaches\citep{zheng2024towards}.
However, the scarcity of high-quality data constrains the scalable training of VLN at the level of fine-grained atomic joint actions~\citep{song2025towards}.
Consequently, most contemporary VLN frameworks output either discrete directions or continuous velocity commands, relying on pre-designed low-level locomotion models for final execution~\citep{li2025urbanvla}.
This architectural decoupling obstructs the synergy between high-level perceptive planning and low-level motor control, hindering the exploration of globally optimal solutions~\citep{he2024agile}.
Furthermore, locomotion models based on the velocity-command-tracking paradigm lack sufficient flexibility to support rapid velocity fluctuations and agile gait adjustments~\citep{rudin2022advanced}.  
Such dual limitations prevent existing VLN methods from reconciling agile obstacle avoidance with robust traversal in dynamic, unstructured environments~\citep{wang2025rethinking}.

\begin{figure}[t]
\centerline{\includegraphics[width=8.7cm]{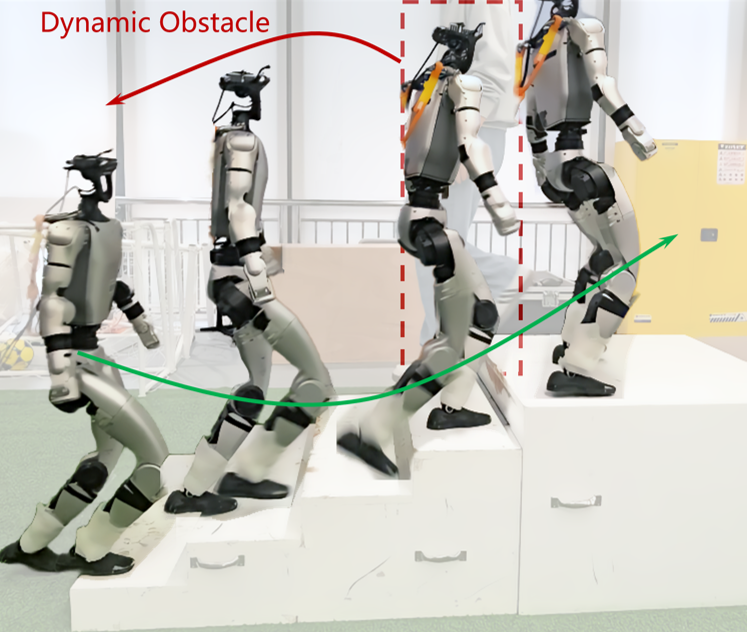}}
\caption{Snapshots of dynamic obstacle avoidance on stairs.
FocusNav anchors perception to navigational intent and adaptively maintains motion stability, significantly enhancing navigational performance in dynamic and complex scenarios.}
\label{fig1}
\end{figure}

\begin{figure}[t]
\centerline{\includegraphics[width=8.7cm]{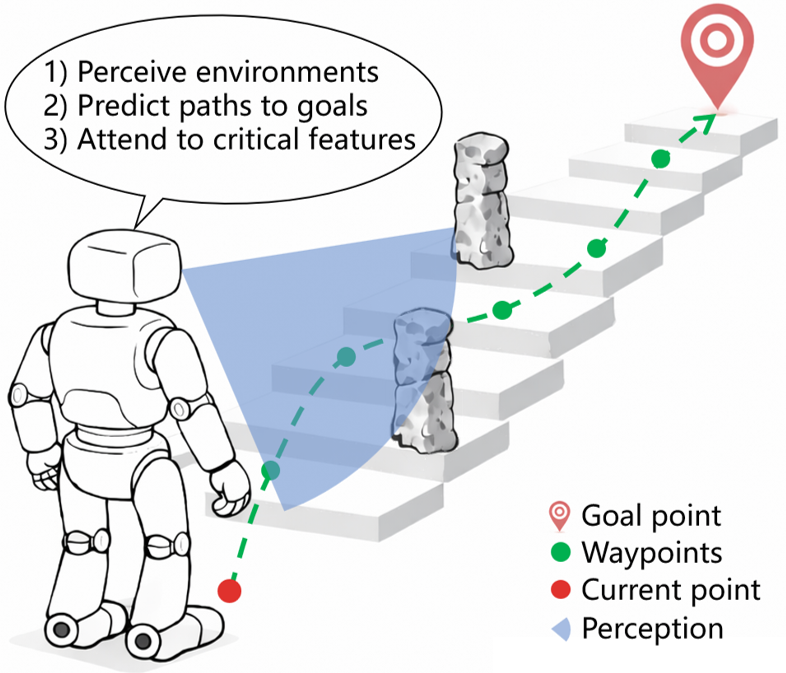}}
\caption{Schematic of the PPA neural circuit paradigm.
The system extracts environmental features through perception and predicts waypoints along the path by integrating the target point with the current location.
An attention mechanism is then employed to focus on critical environmental information, enhancing mobile navigation performance.}
\label{fig2}
\end{figure}

Drawing inspiration from the biological ``cerebrum-cerebellum'' coordination mechanism~\citep{tan2025roboos}, the cerebrum decomposes long-range navigation tasks into sequences of local waypoints through spatial reasoning, while the cerebellum coordinates whole-body movements via reflexive neural circuits to achieve local waypoint tracking~\citep{xiao2025static}.
Fig.~\ref{fig2} visualizes the operational mechanism of neural circuits in biological navigation, referred to as the Perception-Prediction-Attention (PPA) paradigm~\citep{sheikder2025bioinspired}.
Specifically, once a local waypoint is determined, the system extracts feature information by observing the surrounding environment (Perception), pre-simulates potential trajectories in conjunction with target waypoints (Prediction), and dynamically and selectively focuses on critical environmental features along the heading direction (Attention)~\citep{sanyal2025real}.
This paradigm enables low-latency responses for tasks such as terrain adaptation and dynamic obstacle avoidance, ultimately achieving robust local navigation~\citep{de2025bio}.

In this work, we propose FocusNav, a spatial selective attention framework designed to enhance the locomotion stability and navigational efficiency of humanoid robots in complex environments.
FocusNav employs a Waypoint-Guided Spatial Cross-Attention (WGSCA) mechanism that utilizes predicted collision-free waypoints to anchor the perception process, ensuring the model aggregates environmental features directly aligned with the intended motion trajectory.
To further mitigate the risks posed by precarious terrains, we introduce a Stability-Aware Selective Gating (SASG) module.
This module dynamically modulates the robot's perception field by truncating distal environmental information when the robot's stability decreases, thereby forcing the policy to prioritize immediate foothold safety.
We conduct extensive experiments on the Unitree G1 humanoid robot in a variety of challenging scenarios.
The results demonstrate that FocusNav significantly outperforms baseline methods in navigation success rate, collision avoidance efficiency, and motion stability.

In summary, the contributions of this work are as follows:
\begin{itemize}
\item A Waypoint-Guided Spatial Cross-Attention mechanism that aligns navigational intent with environmental context, ensuring that feature aggregation is spatially anchored to the predicted trajectory.
\item A Stability-Aware Selective Gating module that adaptively modulates the perception field based on the robot's dynamic stability, prioritizing local terrain features to ensure motion stability.
\item Extensive experimental results on real robots demonstrate that FocusNav significantly enhances navigation success rate, collision avoidance capability, and motion stability in dynamic and complex environments.
\end{itemize}

\section{Related Works}\label{Related}
\subsection{Humanoid Perceptive Locomotion}\label{Related-1}
In recent years, significant progress has been made in the perceptive locomotion of humanoid robots concerning complex terrain traversal and obstacle avoidance~\citep{sun2025dpl, sun2025learning}.
Existing research predominantly utilizes onboard LiDAR and RGB-D sensors to achieve environmental perception through elevation maps, depth maps, or point clouds~\citep{wang2025beamdojo}.
However, these methods still encounter technical constraints in practical applications~\citep{ren2025vb}.
On one hand, the processing of elevation and depth maps is highly susceptible to computational latency and environmental noise, respectively, which compromises terrain perception accuracy and responsiveness to dynamic obstacles~\citep{wang2025omni}.
On the other hand, the prevailing velocity-command-tracking paradigm limits the flexibility of gait adjustments and introduces an inherent objective conflict between obstacle avoidance reward terms and velocity tracking tasks during training, making it difficult to achieve globally optimal motion~\citep{he2025attention}.
Although some studies have attempted to incorporate target position tracking, they remain confined to static obstacle avoidance in flat environments~\citep{ben2025gallant}.
Simultaneously achieving dynamic obstacle avoidance and robust traversal in unstructured environments, particularly in coordinating terrain information with obstacle constraints to prevent motion conflicts, remains a critical and unresolved challenge.

\subsection{Local Navigation in Complex Environments}\label{Related-2}
Local navigation systems aim to achieve real-time path planning, dynamic obstacle avoidance, and robust locomotion from a robot’s current position to the subsequent local waypoint~\citep{hoeller2024anymal}.
Early research primarily employed modular pipelines, integrating high-level perceptive planning with low-level motor skills, particularly on quadrupedal platforms~\citep{escontrela2025gaussgym}.
To enhance terrain traversability assessment, hierarchical learning frameworks were introduced to jointly train navigation and locomotion policies, fostering a deeper understanding of robot-environment interactions~\citep{roth2025learned}.
While end-to-end visual navigation strategies effectively mitigate system latency and cumulative errors, the inherent flexibility of locomotion models remains constrained by the prevailing velocity-command-tracking paradigm~\citep{cai2025navdp}.
Recent advancements have significantly improved obstacle avoidance and traversal performance in complex terrains by incorporating local waypoint tracking~\citep{lin2025let} and utilizing cross-attention mechanisms to strengthen the capability of motion policies in extracting critical perceptive features~\citep{yang2025spatially}.
However, fully leveraging rich environmental visual information to maintain navigational stability during simultaneous terrain adaptation and dynamic obstacle avoidance remains a pivotal challenge~\citep{zhu2025vr}.
Distinct from existing methodologies, this paper draws inspiration from biological principles to construct a local navigation framework for humanoid robots based on the PPA mechanism.
By guiding the robot to focus on task-relevant key perceptive information, our approach achieves robust mobile navigation performance in extremely complex scenarios.

\section{GuideOracle: Learning Oracle Policy for Waypoint Tracking}\label{GuideOracle}
To bridge the gap between rich environmental information acquired from high-level perception and low-level motion execution, we first develop GuideOracle, a privileged policy specifically engineered for local waypoint tracking.
Independent of visual perception, GuideOracle leverages the full observability of the simulation environment to directly access precise local maps and target coordinates.
This enables the achievement of highly robust waypoint tracking, thereby providing optimal supervisory signals for the subsequent training of our vision-based navigation policy.

\subsection{Problem Formulation}\label{GuideOracle-1}
We formulate the learning of the GuideOracle as a Markov Decision Process (MDP), defined by the tuple $\mathcal{M}=(\mathcal{S},\mathcal{A},\mathcal{P},\mathcal{R},\gamma)$.
The $\mathcal{S}$ and $\mathcal{A}$ represent the state and action spaces, respectively.
The system's transition dynamics are represented by $\mathcal{P}(s_{t+1} | s_t, a_t)$, $\mathcal{R}$ is the reward function, and $\gamma\in[0,1)$ is the discount factor.
The objective of the GuideOracle is to optimize the policy $\pi(a_t | s_t)$ to maximize the expected cumulative discounted reward:
\begin{equation}\label{eq1}
\max_{\pi} J(\pi) = \mathbb{E} \left[ \sum_{t=0}^{\infty} \gamma^t r(s_t, a_t) \right].
\end{equation}
Consider a three-dimensional environment $E\subset\mathbb{R}^{3}$, the robot follows a policy $\pi$ to reach a target region $\mathcal{G}\subset\mathbb{R}^{3}$ within a finite time horizon $t \leq T_{\mathrm{max}}$, such that the relative target position satisfies $\Vert p_{t}\Vert<\epsilon$, where $\epsilon>0$ represents a specified tolerance.
We employ Isaac Gym~\citep{makoviychuk2021isaac} for simulation and Proximal Policy Optimization (PPO)~\citep{schulman2017proximal} to train the GuideOracle.

\subsection{Observation and Action Space}\label{GuideOracle-2}
The observation space includes the base linear and angular velocities, joint positions and velocities, previous actions, navigation commands, and terrain information sampled around the robot.
Specifically, the navigation command $c_{t} \in \mathbb{R}^{4}$ is defined as the $X$-$Y$ axis vector position $p_{\mathrm{xy}}$ of the robot relative to the target position, the angular deviation $e_{\mathrm{yaw}}$ between the robot’s $X$-axis direction and $p_{\mathrm{xy}}$, and the remaining time to reach the target position.
The navigation goal is established at the initial time $t_{0}$ of each episode as the relative target position sampled within the robot's local coordinate frame and the maximum completion time $T_{\mathrm{max}}$.
Unlike previous works~\citep{lin2025let} that rely solely on elevation maps $I_{e}\in \mathbb{R}^{N_{t}}$, where $N_{t}=H_{t}\times W_{t}$, we additionally include a traversability map $I_{t}\in \mathbb{R}^{N_{t}}$, which provides explicit encoding of obstacle regions in the environment.
The actions are interpreted as target joint positions given to the motors where a PD controller transforms them into joint torques.

\begin{figure}[t]
\centerline{\includegraphics[width=8.7cm]{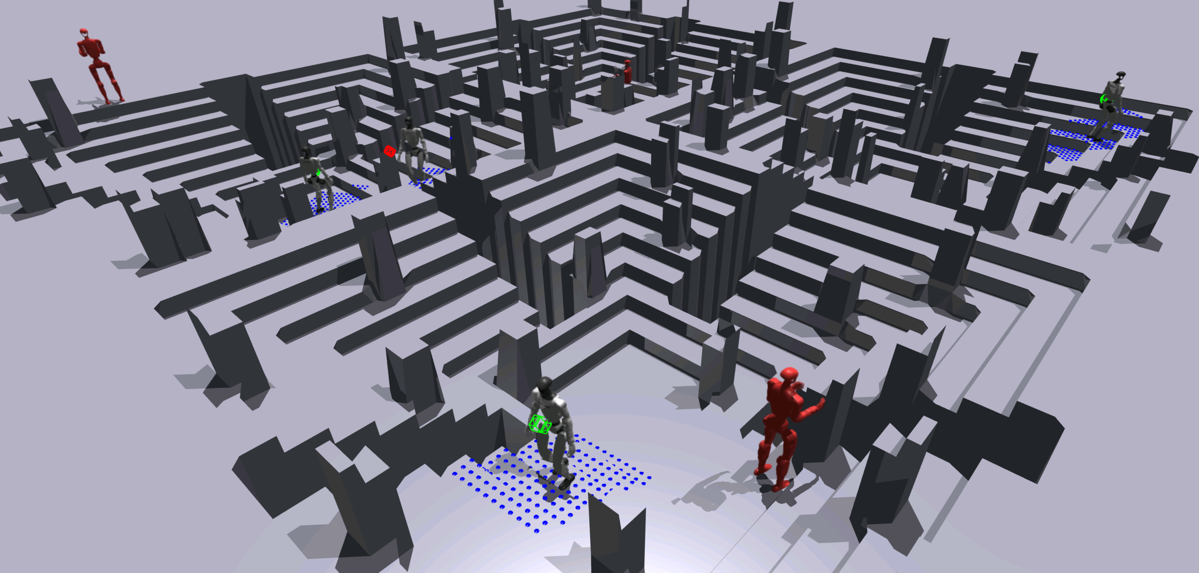}}
\caption{The terrain used for simulation training includes rugged terrain such as stairs, slopes, and gaps.
The environment features dense static obstacles such as forest-like pillars, along with dynamic obstacles represented by red robots following randomized trajectories.}
\label{fig3}
\end{figure}

\begin{figure*}[t]
\centerline{\includegraphics[width=18cm]{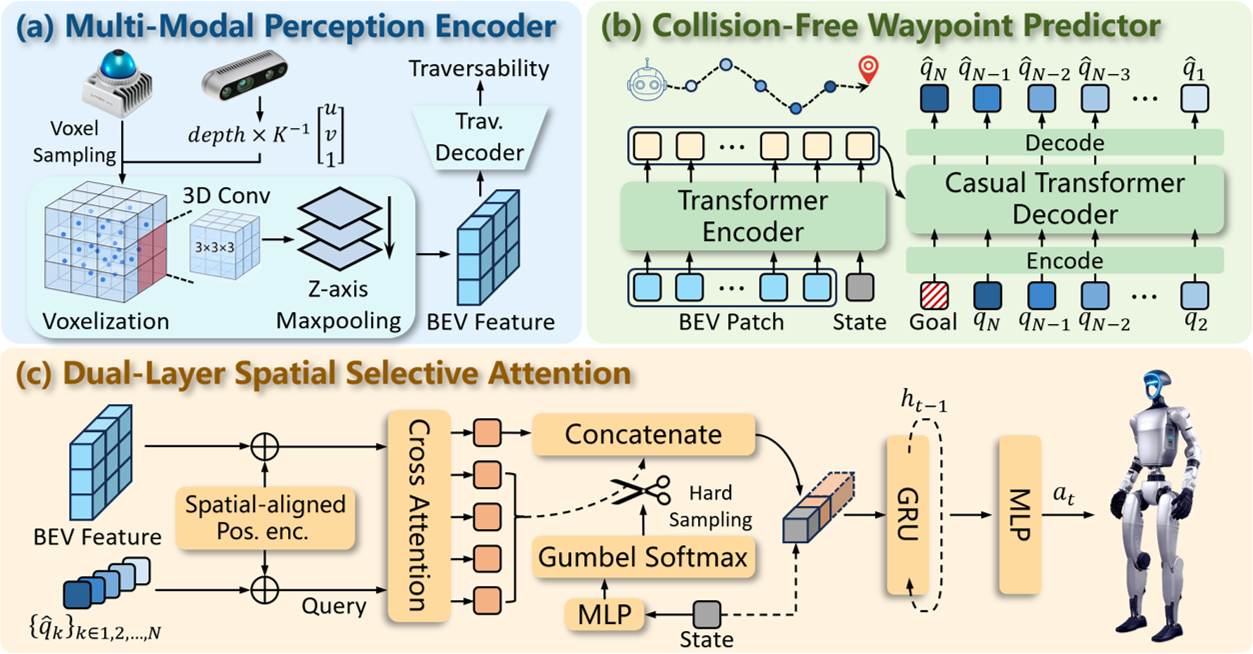}}
\caption{Overview of the FocusNav framework.
(a) Multi-modal perception encoder fuses spatially aligned LiDAR and depth camera data, utilizing voxel feature encoding and convolutional feature extraction to generate BEV feature of the environment.
(b) Collision-free waypoint predictor encodes BEV feature and the robot’s proprioceptive states.
Combined with the navigation goal, a goal-conditioned autoregressive decoder performs backward prediction of trajectory waypoints, enhancing the rationality of the trajectory.
(c) Dual-layer spatial selective attention employs cross-attention, using predicted waypoints to guide the model’s focus toward high-relevance features along the planned trajectory.
Furthermore, a Gumbel Softmax-based gating mechanism dynamically shifts the focus toward local terrain features beneath the robot's feet according to its stability.}
\label{fig4}
\end{figure*}

\subsection{Rewards and Terrains Design}\label{GuideOracle-3}
The reward follows Ben et al.~\citep{ben2025gallant} with velocity tracking reward replaced by target position tracking reward:
\begin{equation}\label{eq2}
r_{\mathrm{task}}=\frac{1}{1+\Vert p_{\mathrm{xy}}\Vert^{2}} \cdot \frac{\mathds{1}(t>T_{\mathrm{max}}-T_{r})}{T_{r}},
\end{equation}
where $T_{r}$ is a parameter defining the duration of the reward, which varies with the sampled $T_{\mathrm{max}}$.
Additionally, we introduce an orientation-tracking reward:
\begin{equation}\label{eq3}
r_{\mathrm{ori}}=exp(-\Vert e_{\mathrm{yaw}}-\omega_{\mathrm{yaw}}\Vert\times\delta_{\mathrm{yaw}}),
\end{equation}
where $\omega_{\mathrm{yaw}}$ denotes the yaw angular velocity of the robot base and $\delta_{\mathrm{yaw}}$ is the weighting coefficient.
This reward encourages the robot to align its heading with the target, leveraging the superior stability of forward locomotion on complex terrain.
By prioritizing $X$-axis movement, the policy restricts lateral and backward maneuvers to essential emergency avoidance, ensuring both navigational efficiency and gait robustness.

To enable robots to learn robust navigation strategies in complex scenarios via simulation, we construct a diverse training environment integrating unstructured rugged terrains and high-density dynamic spaces, as shown in Fig.~\ref{fig3}.
This environment incorporates unstructured rugged features including stairs, slopes, and gaps, so as to replicate the locomotion challenges caused by terrain elevation changes and traversal constraints in real-world scenarios.
In addition, we simulate the complex spatiotemporal uncertainties of real environments by integrating forest-like scenes composed of dense pillars, crowd-dense scenarios, and randomly moving robot entities as dynamic obstacles.
This multi-dimensional environment design ensures that the learned policies exhibit excellent generalization performance when confronted with various spatial constraints and dynamic disturbances.
Based on obstacles and terrain gaps, the environmental information acquired by the robot includes elevation maps and traversability map $I_{\mathrm{et}}=\{I_{e},I_{t}\}$.
Traversability map explicitly characterizes the non-traversability of obstacle and gap regions.
In addition, the randomly sampled navigation target positions are processed by a reachability optimization layer, which projects the target positions located in non-traversable regions onto the nearest reachable areas along the direction connecting the robot and the target, thereby ensuring the effectiveness of the robot's navigation tasks in complex environments.

\section{FocusNav: Local Navigation with Spatial Selective Attention}\label{FocusNav}
In this section, we present a detailed exposition of the FocusNav framework, which is engineered to enhance the local navigation performance of humanoid robots in complex and dynamic environments by emulating the PPA paradigm in biological neural navigation circuits.
As illustrated in Fig.~\ref{fig4}, the FocusNav comprises three synergistically integrated modules: a multi-modal perception encoder for comprehensive environmental feature extraction, a collision-free waypoint predictor for safe path planning, and a dual-layer spatial selective attention mechanism for prioritized feature processing.
By mimicking the hierarchical information processing of biological systems, FocusNav enables the robot to precisely focus on critical environmental cues, achieving effective dynamic obstacle avoidance while maintaining superior motion stability during goal-directed navigation tasks.

\subsection{Multi-Modal Perception Encoder}\label{FocusNav-1}
\subsubsection{Efficient Sensor Simulation}\label{FocusNav-1-1}
To acquire environmental information, we utilize LiDAR and depth camera commonly integrated into robotic platforms.
We implement high-performance LiDAR rendering~\citep{wang2025omni} and depth synthesis~\citep{sun2025dpl} using NVIDIA Warp~\citep{macklin2022warp} for GPU-accelerated raycasting.
To enhance the speed of dynamic geometry rendering, we adopt the method proposed by Ben et al.~\citep{ben2025gallant} to precompute the Bounding Volume Hierarchy (BVH) for meshes within its local frames.
At each simulation step, the ray origin $p_{r}$ and direction $d_{r}$ are transformed into the local frame.
Ray-mesh intersections are then efficiently computed using the precomputed BVH, after which the resulting local intersection coordinates are transformed back into the world frame.
The raycasting process can be formally represented as:
\begin{equation}\label{eq4}
\mathrm{raycast}(M, T, p_{r}, d_{r}) = T \cdot \mathrm{raycast}(M, T^{-1}p_{r}, R^{-1}d_{r}),
\end{equation}
where $M$ denotes the mesh geometry, $T$ represents the homogeneous transformation matrix, and $R$ indicates the rotational component.

\subsubsection{Cross-Modal Spatial Alignment}\label{FocusNav-1-2}
\begin{figure}[t]
\centering  
\subfigure[]{
\label{fig5-1}
\includegraphics[width=4.2cm]{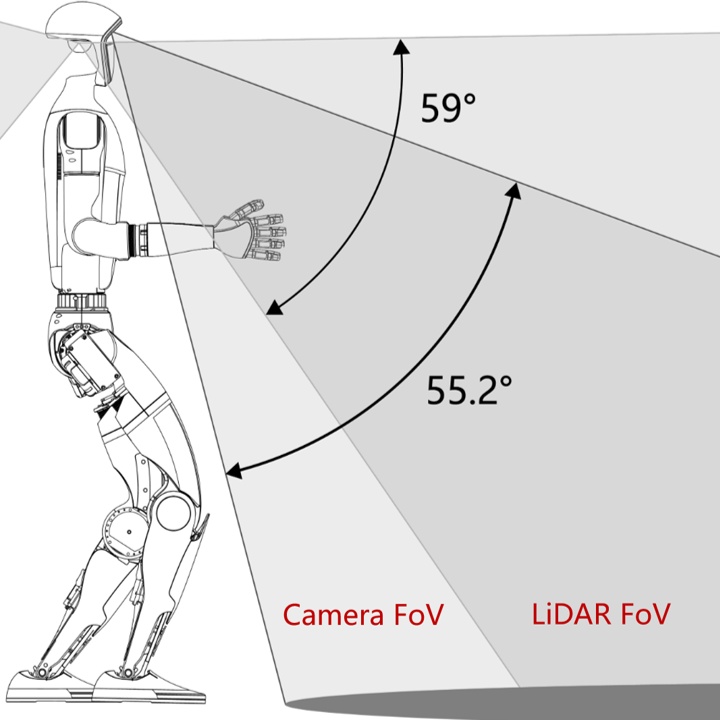}}
\subfigure[]{
\label{fig5-2}
\includegraphics[width=4.2cm]{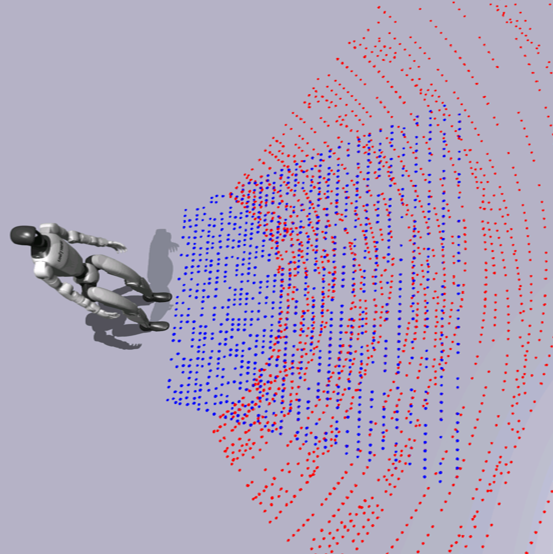}}
\caption{FoV of onboard LiDAR and depth camera for the robot.
(a) Complementary sensing characteristics~\citep{unitree_g1_2026}: The LiDAR is characterized by an extended effective measurement range, whereas the depth camera reliably covers the proximal terrain surrounding the robot's feet.
(b) Cross-modal point cloud in simulation: The red and blue point clouds are from the LiDAR and the depth camera, respectively.}
\label{fig5}
\end{figure}

Owing to the discrepancies in mounting positions and Fields of View (FOV) among various sensors, multi-modal fusion significantly extends the perceptual range and enhances sensing accuracy.
Taking the Unitree G1 humanoid robot~\citep{example_website} as an example, the onboard LiDAR provides a significantly greater observation range than the depth camera, yet it is constrained by a blind spot regarding the immediate terrain surrounding the robot's feet, as illustrated in Fig.~\ref{fig5-1}.
Uniform voxel sampling~\citep{rusu20113d} is employed to extract the sparse point cloud $P_{L}$ from the dense LiDAR data.
Meanwhile, the depth map $D_{C}\in \mathbb{R}^{H \times W}$ is back-projected into the camera-centric point cloud $P_{C}$ using the camera intrinsic matrix $K \in \mathbb{R}^{3 \times 3}$.
For a pixel coordinate $(u, v)$ with depth value $d$, the corresponding 3D point $p_{c} \in P_{C}$ is computed as:
\begin{equation}\label{eq5}
p_{c} = d \cdot K^{-1} [u, v, 1]^T,
\end{equation}
then $P_{C}=\{p_{c,i}\}_{i=1}^{N_{c}}$, where $N_{c}=h\times w$ represents the number of point clouds corresponding to the image.
By leveraging sensor extrinsic parameters, both $P_{L}$ and $P_{C}$ are transformed into the world frame and concatenated to form the cross-modal point cloud $P_{\mathrm{LC}}=\{P_{L}^{w}, P_{C}^{w}\}$.
As demonstrated in Fig.~\ref{fig5-2}, the red points represent the LiDAR-derived point cloud $P_{L}$, while the blue points denote the point cloud $P_{C}$ acquired via the depth camera.

\subsubsection{BEV Feature Extraction}\label{FocusNav-1-3}
Inspired by the VoxelNet architecture~\citep{zhou2018voxelnet}, we develop a lightweight perception encoder to transform the unstructured point cloud $P_{\mathrm{LC}}$ into a structured Bird's-Eye-View (BEV) representation.
The process begins with voxelization $\mathcal{V}(\cdot)$, where $P_{\mathrm{LC}}$ is discretized into a 3D grid of size $D \times H \times W$.
A voxel feature encoding layer learns a high-dimensional latent representation for each non-empty voxel, effectively capturing local geometric primitives.
The encoded 4D tensor is processed through 3D convolution $\mathcal{C}_{\mathrm{3D}}$.
The resulting feature volume is subsequently projected onto the BEV plane via a $Z$-axis global maximum pooling operator $\mathcal{P}_{z}$.
A multi-scale convolutional network $\mathcal{G}$ is utilized to perform feature fusion, yielding the spatially consistent BEV feature $F_{\mathrm{bev}} \in \mathbb{R}^{C \times H \times W}$.
Formally, this composite mapping is expressed as:
\begin{equation}\label{eq6}
F_{\mathrm{bev}} = \mathcal{G} \circ \mathcal{P}_{z} \circ \mathcal{C}_{\mathrm{3D}} \circ \mathcal{V}(P_{\mathrm{LC}}),
\end{equation}
where $\circ$ denotes the function composition. 
This pipeline efficiently extracts a semantic-rich and spatially consistent environment representation.
To augment latent features with semantic traversability, a traversability decoder upsamples the $F_{\mathrm{bev}}$ into a traversability map $\hat{I}_{t} \in \mathbb{R}^{H_{t} \times W_{t}}$.
The network is optimized via the Binary Cross-Entropy loss:
\begin{equation}\label{eq7}
\mathcal{L}_{t} = - \frac{1}{N_{t}} \sum_{i=1}^{N_{t}}\left[I_{t,i} \log(\hat{I}_{t,i}) + (1 - I_{t,i}) \log(1 - \hat{I}_{t,i}) \right],
\end{equation}
where $I_{t,i} \in \{0, 1\}$ represents the ground-truth traversability.

\subsection{Collision-Free Waypoint Predictor}\label{FocusNav-2}
\subsubsection{Backward Prediction Paradigm}\label{FocusNav-2-1}
To bridge the gap between high-level perception and low-level control, we propose a goal-conditioned backward prediction paradigm for generating collision-free navigation waypoints.
Unlike conventional forward prediction paradigms that are prone to accumulating compound errors~\citep{liu2025efficient}, backward prediction paradigm explicitly generates waypoints starting from the navigation target $g_{n}$ back to the robot's current position.
This formulation enforces a reverse causal dependency among waypoints, yielding a goal-conditioned reasoning chain that ensures global-to-local consistency~\citep{zhou2025act2goal}.
By anchoring the trajectory to the navigation target, we enhance the logical coherence and traversability of the planned path.
Formally, given the observation $\mathcal{O}$, which includes the BEV features $F_{\mathrm{bev}}$, the robot’s proprioceptive state $S_{p}$ and the navigation target $g_{n}$, the predictor models the trajectory distribution as a joint probability of waypoints $\hat{q}$ in reverse chronological order:
\begin{equation}\label{eq8}
\begin{split}
p(\hat{q}_{1:N} &\mid \mathcal{O}) = p(\hat{q}_{N} \mid \mathcal{O}) \cdot p(\hat{q}_{N-1} \mid \mathcal{O}, \hat{q}_{N})\\
&\cdot p(\hat{q}_{N-2} \mid \mathcal{O}, \hat{q}_{N-1:N})\cdot \dots \cdot p(\hat{q}_{1} \mid \mathcal{O}, \hat{q}_{2:N}).
\end{split}
\end{equation}
This backward chaining mechanism utilizes the navigation target $g_{n}$ as a strong prior to regularize the entire decoding sequence, forcing the latent representation to focus on the reachable manifold of the terrain, thereby enhancing the logical consistency and robustness of trajectory prediction in complex unstructured environments.

\subsubsection{Autoregressive Network with Latent Consistency}\label{FocusNav-2-2}
To implement the backward prediction paradigm, we develop a goal-conditioned Transformer architecture that recursively predicts waypoints in reverse-time autoregression by integrating multi-modal environmental context with robot proprioceptive states, as shown in Fig.~\ref{fig4}(b).
The transformer encoder serves as a perception backbone that extracts context-aware latent embeddings by processing a concatenated sequence of flattened BEV patches and the robot’s proprioceptive state $S_{p}$.
Utilizing the causal transformer decoder, we generate the waypoint sequence $\{\hat{q}_{k}\}_{k=1}^{N}$ autoregressively starting from the navigation target $g_n$.
This setup ensures that each waypoint prediction is strictly conditioned on both the global goal and the sequence of previously predicted distal points.
The network is optimized via a multi-task objective function that jointly supervises spatial accuracy and temporal coherence.
Specifically, we employ a waypoint reconstruction loss $\mathcal{L}_{\mathrm{rec}}$ based on Mean Squared Error (MSE) to minimize the geometric discrepancy between predicted coordinates $\hat{q}_{k}$ and ground-truth waypoints $q_{k}$.
To address the temporal inconsistency and error propagation inherent in continuous latent space autoregression~\citep{zhang2025chain}, we introduce latent consistency regularization $\mathcal{L}_{\mathrm{reg}}$.
This mechanism explicitly aligns the predicted latent tokens $\hat{x}_k$ with the expert latent representations $f_{\mathrm{enc}}(q_k)$, forcing the model to adhere to the feasible manifold.
The waypoint prediction loss $\mathcal{L}_{p}=\mathcal{L}_{\mathrm{rec}}+\mathcal{L}_{\mathrm{reg}}$ is formulated as:
\begin{equation}\label{eq9}
\mathcal{L}_{p} = \sum_{k=1}^{N} \Vert \hat{q}_k - q_k \Vert ^{2} + \lambda_{\mathrm{reg}} \sum_{k=1}^{N} \Vert \hat{x}_k - f_{\mathrm{enc}}(q_k) \Vert ^{2},
\end{equation}
where $N$ represents the number of waypoints, $\lambda_{\mathrm{reg}}$ is a weighting coefficient used to balance waypoints prediction accuracy and temporal consistency.

\subsection{Dual-Layer Spatial Selective Attention}\label{FocusNav-3}
To extract task-critical features from high-dimensional environment representations, we propose a dual-layer spatial selective attention module, as illustrated in Fig.~\ref{fig4}(c).
This module mimics selective attention by dynamically focusing on navigation-relevant spatial cues, thereby enhancing the control robustness of humanoid robots in unstructured terrains~\citep{matthis2018gaze}.

\subsubsection{Waypoint-Guided Spatial Cross-Attention}\label{FocusNav-3-1}
The first layer of our attention mechanism utilizes the predicted collision-free waypoints $\{\hat{q}_{k}\}_{k=1}^{N}$ to guide the perception process. 
Specifically, we utilize the latent tokens $\{\hat{x}_k\}_{k=1}^{N}$, the intermediate representations of the predicted waypoints, as queries, while the sequence of flattened BEV patches serves as keys and values.
To reinforce the spatial correspondence between these modalities, we introduce a Spatially-Aligned Position Embedding (PE) mechanism.
By projecting both the waypoints and BEV patches into a unified coordinate system, we employ sinusoidal position embeddings to preserve their relative spatial topological structure.
Consequently, the cross-attention module performs precise point-wise attention over the environment, ensuring that the attention scores accurately reflect the physical proximity between the robot's intended path and surrounding terrain features.
Formally, this process yields a sequence of waypoint-focused map embeddings $\{m_{k}\}_{k=1}^{N} \in \mathbb{R}^{N \times d}$:
\begin{equation}\label{eq10}
m_{k} = \mathrm{Attn}(\hat{x}_k+PE(\hat{q}_{k}), F_{\mathrm{bev}}+PE(F_{\mathrm{bev}}), F_{\mathrm{bev}}),
\end{equation}
where each embedding $m_{k} \in \mathbb{R}^{d}$ represents the environmental feature context specifically aggregated at the spatial location of the $k$-th predicted waypoint.
This operation effectively anchors the perception to the planned path, filtering out irrelevant background noise and focusing the representation on the local traversability and obstacles critical to the robot's immediate movement.

\subsubsection{Stability-Aware Selective Gating}\label{FocusNav-3-2}
To further ensure the robot's immediate survival in highly dynamic or precarious terrains, we introduce a gating mechanism as the second layer of our spatial attention module.
This layer operates on the trajectory-aligned map embeddings $\{m_{k}\}_{k=1}^{N}$ by dynamically truncating distal spatial information based on the robot’s proprioceptive state $S_{p}$.
Specifically, $S_{p}$ is processed through a Multi-Layer Perceptron (MLP) to generate the gating logits $V^{g} \in \mathbb{R}^2$, representing the activation and deactivation probabilities for the gating module.
To maintain end-to-end differentiability while performing discrete selection, the gate control variable $g$ is sampled using the Gumbel-Softmax distribution~\citep{jang2016categorical}:
\begin{equation}\label{eq11}
g = \mathrm{GumbelSoftmax}(V^{g}, \tau) \in \{0, 1\},
\end{equation}
where $\tau$ denotes the temperature parameter.
The variable $g$ serves as a binary switch that governs the inclusion of distal environmental context $\{m_{k}\}_{k=2}^{N}$.
Specifically, when $g=1$, the activated gate indicates a stable robotic state, whereby the model aggregates environmental information along the entire trajectory to fulfill the navigation task.
Conversely, $g=0$ signifies a decrease in dynamic stability, prompting the model to truncate far-field environmental embeddings and refocus exclusively on the immediate terrain features surrounding the robot’s feet.
The resulting stability-aware hybrid map embedding $m^h$ is formulated as:
\begin{equation}\label{eq12}
m^{h} = m_{1} + g \cdot \sum_{k=2}^{N} m_{k}.
\end{equation}
By prioritizing local foothold safety over distant navigation targets during periods of instability, this mechanism ensures effective recovery and robust traversal in complex, non-structured environments.
To supervise the gating mechanism, we derive a stability metric $\mathcal{S}_{m} \in [0, 1]$ based on the robot's real-time proprioceptive feedback, including angular velocities $\omega$ and Euler angles $\phi$.
The stability metric is formulated to quantify the deviation from a balanced state:
\begin{equation}\label{eq13}
\mathcal{S}_{m} = \frac{1}{1 + k_{1}|\phi_{\mathrm{xy}}|^2 + k_{2}|\omega_{\mathrm{xy}}|^2},
\end{equation}
where $\{k_i\}$ are scaling constants.
The gating module is optimized via a cross-entropy loss $\mathcal{L}_{g}$ that aligns the gating probabilities $p_k = P(g=k)$ with the stability metric $\mathcal{S}_m$:
\begin{equation}\label{eq14}
\mathcal{L}_{g} = -\mathbb{E} \left[ \mathcal{S}_{m} \log p_1 + (1-\mathcal{S}_{m}) \log p_0 \right].
\end{equation}
This objective ensures that the gating module learns to autonomously modulate its perceptual focus based on the robot's dynamic equilibrium.

The selectively extracted hybrid map embedding $m^{h}$ is concatenated with the robot's state $S_{p}$ and processed by a GRU-based control policy.
By leveraging its hidden state $h_t$, the GRU maintains a latent memory of the environment, ensuring that global navigation consistency is preserved even when the gating mechanism temporarily shifts focus toward immediate local footholds.
The control policy is optimized using a Behavior Cloning (BC) objective $\mathcal{L}_{\mathrm{bc}}$ to minimize the discrepancy between the FocusNav's actions and the GuideOracle’s expert demonstrations.
To ensure that the internal representations are task-relevant, the entire pipeline is trained in an end-to-end differentiable manner. 
The total training loss $\mathcal{L}_{\mathrm{total}}$ can be expressed as:
\begin{equation}\label{eq15}
\mathcal{L}_{\mathrm{total}} = \mathcal{L}_{\mathrm{bc}} + \lambda_{1}\mathcal{L}_{t} + \lambda_{2}\mathcal{L}_{p} + \lambda_{3}\mathcal{L}_{g},
\end{equation}
where the coefficients $\{\lambda_{i}\}_{i=1,2,3}$ are hyperparameters used to balance the contribution of each auxiliary task.
This joint optimization ensures that the perception, prediction, attention, and control modules are mutually aligned for robust mobile navigation in unstructured environments.

\section{Experiments}\label{Experiments}
In this section, we present extensive experimental results of FocusNav conducted in both simulation and real-world environments.
Our experiments are designed to systematically evaluate the effectiveness of the proposed framework in navigating complex and dynamic environments, specifically addressing the following research questions:
(1) Does the WGSCA mechanism effectively leverage predicted waypoints to anchor task-relevant environmental features?
(2) Does the SASG module effectively modulate the attention span to prioritize stability when necessary?
(3) Does FocusNav significantly outperform baseline methods in humanoid local navigation across dynamic and unstructured terrains?

\subsection{Experimental Setup}\label{Experiments-1}
We deploy FocusNav on the Unitree G1, a 29-DoF humanoid robot equipped with a Livox MID-360 LiDAR and a RealSense D435i depth camera for environmental perception.
Training is conducted within the IsaacGym simulation environment~\citep{makoviychuk2021isaac}, distributed across four NVIDIA RTX 4090 GPUs.
For real-world deployment, the robot’s relative localization is estimated via the Fast-LIO algorithm~\citep{xu2021fast} using LiDAR point clouds.

\subsubsection{\textbf{Baselines}}\label{Experiments-1-1}
To evaluate the contribution of each core component in FocusNav, we compare our framework against the following three baseline methods:
\begin{itemize}
\item \textbf{Gallant}~\citep{ben2025gallant}:
Following the standard feature fusion paradigm, this baseline directly concatenates processed environmental features with proprioceptive states before feeding them into the policy network.
\item \textbf{Proprioception-Guided Cross-Attention} (\textbf{PGCA})~\citep{he2025attention}: This variant utilizes the robot's proprioceptive state $S$ as the query to attend to the BEV environmental features.
\item \textbf{WGSCA-Only}: This version incorporates our proposed WGSCA mechanism but excludes the SASG module. It is designed specifically to isolate and evaluate the impact of the stability-aware gating module on motion performance. 
\end{itemize}

\subsubsection{\textbf{Metrics}}\label{Experiments-1-2}
We employ the following four quantitative metrics to evaluate the performance of our model:
\begin{itemize}
\item \textbf{Success Rate} $E_{\mathrm{success}}$: The percentage of episodes in which the robot successfully reaches the target location without terminal failure.
\item \textbf{Traverse Rate} $E_{\mathrm{traverse}}$: The ratio of the actual distance traveled by the robot before termination to the planned travel distance.
\item \textbf{Collision Frequency} $E_{\mathrm{collision}}$: The frequency of collision events between the robot and environmental obstacles per unit time.
\item \textbf{Motion Stability} $E_{\mathrm{stability}}$: The average stability metric $\mathcal{S}_{m}$ of robot motion trajectory.
\end{itemize}

\begin{table*}[t]
\centering
\caption{Quantitative simulation results of FocusNav and baseline methods.
Evaluated across flat and unstructured terrains with static/dynamic obstacles, the results demonstrate the robustness of FocusNav in balancing environmental feature extraction and immediate motion stability.
Bold values indicate the best performance in each experimental setting.} 
\label{table1}
\begin{tabular*}{1.0\textwidth}{l@{\extracolsep{\fill}} cccc cccc}
\toprule
\multirow{2}{*}{\textbf{Method}} & \multicolumn{4}{c}{Flat Terrain} &  \multicolumn{4}{c}{Unstructured Terrain}  \\
\cmidrule(lr){2-5}
\cmidrule(lr){6-9}
& $E_{\mathrm{success}}(\%)\uparrow$ & $E_{\mathrm{traverse}}(\%)\uparrow$ &    $E_{\mathrm{collision}}(\%)\downarrow$ & $E_{\mathrm{stability}}\uparrow$ & $E_{\mathrm{success}}(\%)\uparrow$ & $E_{\mathrm{traverse}}(\%)\uparrow$ &    $E_{\mathrm{collision}}(\%)\downarrow$ & $E_{\mathrm{stability}}\uparrow$\\
\midrule
\multicolumn{9}{l}{\textbf{Static Obstacle Environment}}\\
\cmidrule{1-9}
Gallant & $85.45\pm2.12$ & $88.30\pm3.45$ & $2.65\pm0.32$ & $0.79\pm0.04$ & $60.14\pm3.56$ & $65.55\pm4.12$ & $4.42\pm0.45$ & $0.61\pm0.05$ \\
PGCA & $90.67\pm1.55$ & $92.12\pm2.05$ & $2.24\pm0.15$ & $0.82\pm0.03$ & $71.89\pm2.89$ & $75.34\pm3.21$ & $3.52\pm0.25$ & $0.70\pm0.04$ \\
WGSCA-Only & $95.12\pm0.85$ & $96.45\pm1.15$ & $1.69\pm0.06$ & $0.84\pm0.02$ & $82.23\pm2.01$ & $85.17\pm2.55$ & $2.58\pm0.09$ & $0.73\pm0.05$ \\
\textbf{FocusNav} & $\mathbf{97.54\pm0.35}$ & $\mathbf{98.88\pm0.52}$ & $\mathbf{1.42\pm0.05}$ & $\mathbf{0.88\pm0.02}$ & $\mathbf{91.15\pm0.88}$ & $\mathbf{93.20\pm1.12}$ & $\mathbf{2.08\pm0.08}$ & $\mathbf{0.81\pm0.03}$\\
\midrule
\multicolumn{9}{l}{\textbf{Dynamic Obstacle Environment}}\\
\cmidrule{1-9}
Gallant & $68.23\pm4.15$ & $72.12\pm5.22$ & $8.45\pm0.89$ & $0.67\pm0.05$ & $50.32\pm7.67$ & $53.15\pm8.12$ & $12.55\pm1.25$ & $0.55\pm0.07$ \\
PGCA & $75.45\pm3.25$ & $78.88\pm4.05$ & $6.62\pm0.65$ & $0.72\pm0.04$ & $63.67\pm5.45$ & $66.23\pm5.15$ & $8.94\pm0.88$ & $0.62\pm0.06$ \\
WGSCA-Only & $84.34\pm1.85$ & $88.12\pm2.45$ & $3.88\pm0.22$ & $0.77\pm0.03$ & $74.15\pm3.15$ & $78.78\pm4.25$ & $\mathbf{4.32\pm0.35}$ & $0.68\pm0.05$ \\
\textbf{FocusNav} & $\mathbf{93.45\pm1.02}$ & $\mathbf{94.42\pm1.85}$ & $\mathbf{3.75\pm0.15}$ & $\mathbf{0.84\pm0.06}$ & $\mathbf{87.02\pm1.15}$ & $\mathbf{89.17\pm2.99}$ & $4.56\pm0.21$ & $\mathbf{0.76\pm0.04}$\\
\bottomrule
\end{tabular*}
\end{table*}

\subsection{Simulation Results}\label{Experiments-2}
\subsubsection{Quantitative Results}\label{Experiments-2-1}
We report a detailed quantitative comparison of FocusNav against baseline methods in Table~\ref{table1}, evaluated across four quantitative metrics: success rate $E_{\mathrm{success}}$, traverse rate $E_{\mathrm{traverse}}$, collision frequency $E_{\mathrm{collision}}$, and motion stability $E_{\mathrm{stability}}$.
These benchmarks are conducted in diverse scenarios, including flat and unstructured terrains with both static and dynamic obstacles.
For each configuration, we compute the mean and standard deviation from policies trained with three distinct random seeds, each validated across 100 randomized episodes.
The experimental results lead to the following key conclusions and insights:

\textbf{Superiority of FocusNav in complex environments.}
Quantitative results demonstrate that FocusNav consistently outperforms all baseline methods, particularly as environmental complexity increases.
In the most challenging dynamic and unstructured scenario, FocusNav achieves a success rate $E_{\mathrm{success}}$ of $87.02\%$, significantly surpassing Gallant ($50.32\%$) and PGCA ($63.67\%$).
This substantial margin highlights the limitations of standard environmental feature extraction and proprioception-guided attention mechanisms when handling simultaneous terrain disturbances and moving obstacles.
In contrast, our framework maintains high reliability by anchoring environmental attention to the predicted trajectory while dynamically focusing on local foothold features.

\textbf{Enhanced long-range perception via WGSCA.}
Compared to PGCA, which primarily extracts environmental features around the robot's feet via proprioceptive attention, WGSCA enhances long-range proactive perception by anchoring spatial attention to predicted waypoints.
This approach establishes a direct correlation between distal environmental features and the current motion plan, thereby improving adaptability to complex navigation tasks.
By filtering out irrelevant environmental interference outside the intended trajectory, WGSCA ensures that long-range perception is more task-oriented, enabling more predictive and smoother dynamic obstacle avoidance.

\textbf{Necessity of SASG in dynamic perceptual field modulation.}
In unstructured terrain navigation, when local terrain complexity leads to degraded robot stability, the SASG module autonomously truncates distal information.
This compels the motion policy to concentrate attention on immediate foothold features, thereby enhancing instantaneous motion robustness and facilitating the completion of long-range navigation tasks.
Furthermore, in dynamic unstructured environments, FocusNav exhibits a slightly higher collision frequency compared to the WGSCA-only variant.
This reveals a critical trade-off between terrain traversal and dynamic obstacle avoidance in complex environments.
To maintain motion stability throughout long-range tasks, the SASG module enables the robot to prioritize self-stability over aggressive collision avoidance, preventing catastrophic falls and ensuring task completion.

\begin{figure}[t]
\centering  
\subfigure[]{
\label{fig6-1}
\includegraphics[width=4.0cm]{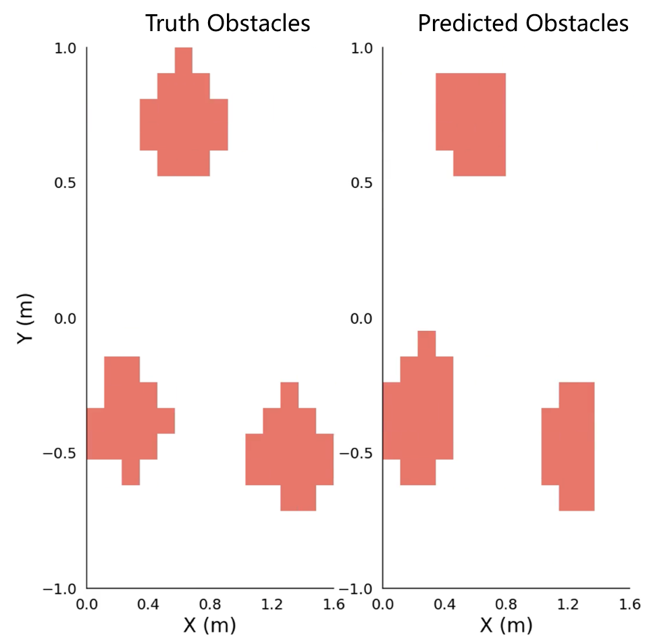}}
\subfigure[]{
\label{fig6-2}
\includegraphics[width=4.35cm]{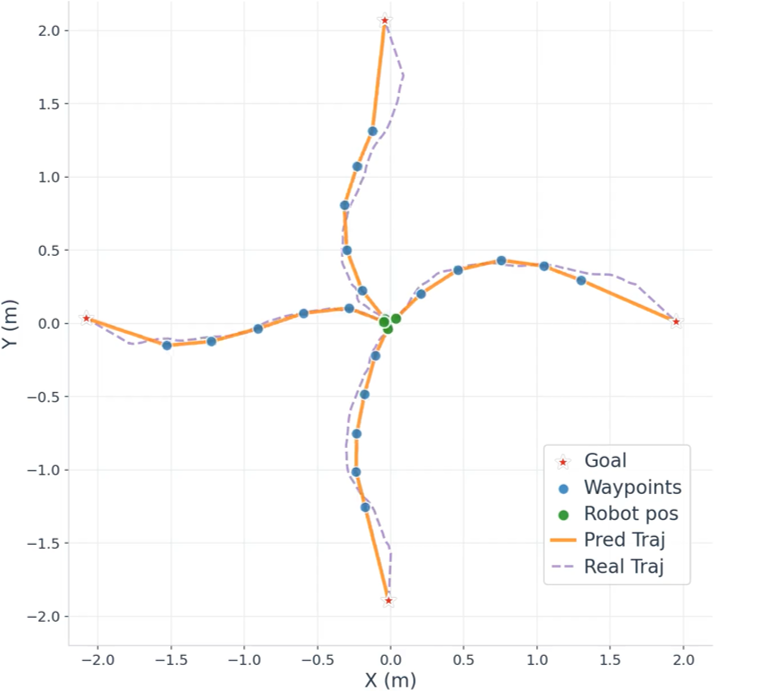}}
\caption{Visualization of traversability and waypoints prediction.
(a) Robot-centric local traversability prediction results.Red regions indicate untraversable areas due to obstacles in the environment.
(b) Waypoint prediction results across four navigation directions.
Collision-free trajectories are predicted within environments featuring diverse obstacles.}
\label{fig6}
\end{figure}

\begin{figure}[t]
\centering  
\subfigure[]{
\label{fig7-1}
\includegraphics[width=4.2cm]{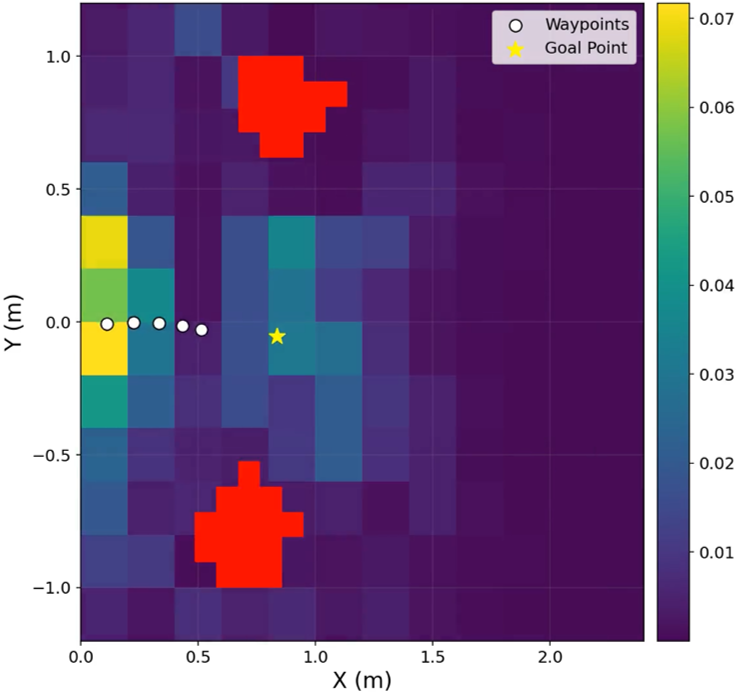}}
\subfigure[]{
\label{fig7-2}
\includegraphics[width=4.2cm]{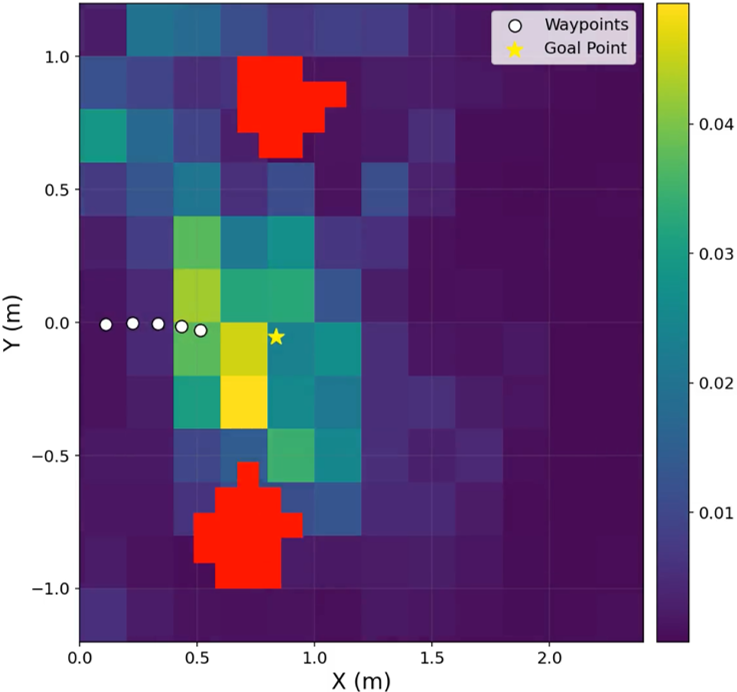}}
\caption{Visualization of waypoint-guided cross-attention over BEV features.
(a) Proximal waypoint: Attention concentrates near the robot's feet, focusing on terrain features critical for foothold safety.
(b) Distal waypoint: Attention anchors to the distant predicted path, extracting long-range environmental cues for navigational consistency.}
\label{fig7}
\end{figure}

\subsubsection{Task-Oriented Perceptual Anchoring via WGSCA}\label{Experiments-2-2}
To further elucidate how WGSCA effectively filters environmental noise and precisely anchors perceptual focus to critical spatial regions, we systematically analyze several key components.
We visualize the terrain traversability reconstructed from the BEV features in a multi-obstacle environment, as shown in Fig.~\ref{fig6-1}, where red regions denote non-traversable areas containing obstacles.
By comparing the reconstructed maps with the ground truth, it is evident that the BEV features effectively identify irregular and disordered obstacles.
Furthermore, we visualize the goal-conditioned backward waypoint trajectory predictions for navigation tasks with different directions of motion,  as shown in Fig.~\ref{fig6-2}. 
The results demonstrate that, by integrating navigational goals with environmental BEV features, the WGSCA module accurately interprets global task intents and generates collision-free waypoints.
This capability provides proactive spatial guidance for long-range navigation tasks.

Furthermore, by comparing the cross-attention distributions guided by waypoints at varying distances, we demonstrate that WGSCA dynamically extracts trajectory-relevant environmental features based on the predicted path.
As illustrated in Fig.~\ref{fig7}, when queried with proximal waypoints, the attention selectively prioritizes the immediate surroundings to extract terrain structures critical for foothold selection.
In contrast, with distal waypoints, attention extends along the anticipated trajectory to distill long-range environmental cues.
This spatial selective attention mechanism enables the model to establish a spatio-temporal correlation between navigational intent and environmental perception, which significantly enhances the consistency of the robot's navigational decision-making in complex and dynamic environments.

\begin{figure}[t]
\centerline{\includegraphics[width=8.7cm]{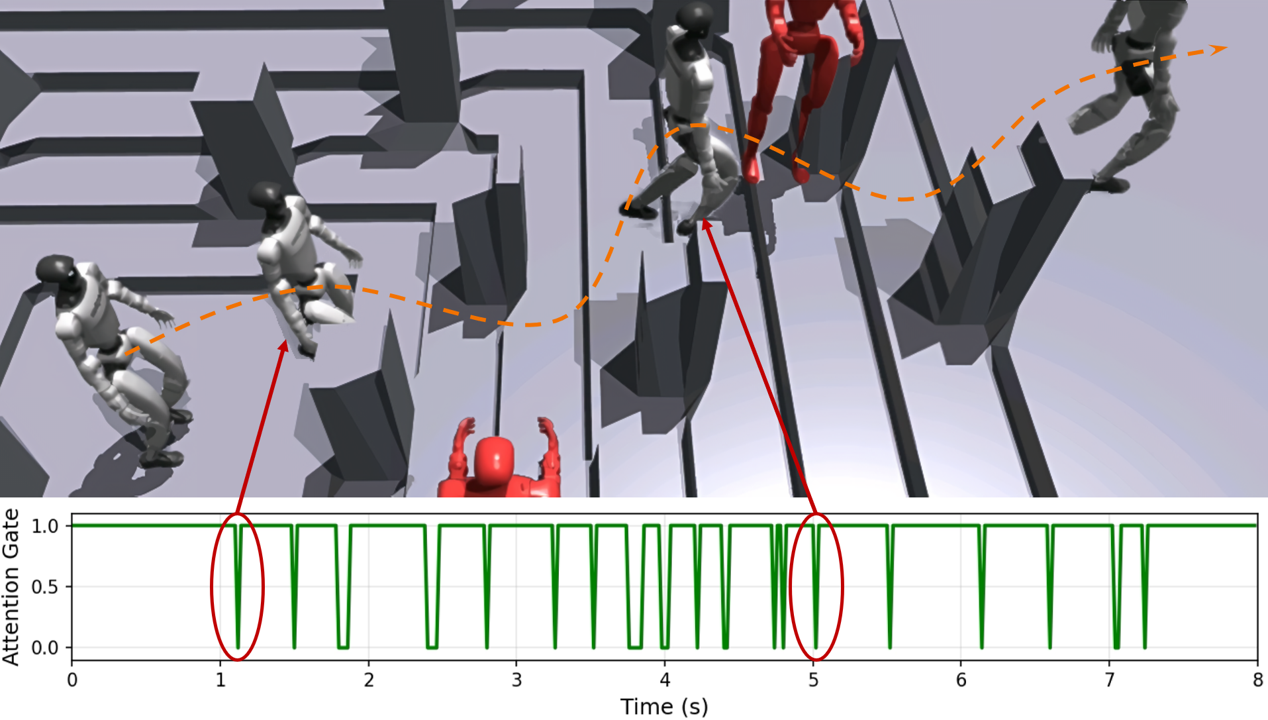}}
\caption{Visualization of attention gates in dynamic and complex terrain navigation.
As the robot transitions from flat ground to stairs, the attention gating selectively truncates distal perceptual information to prioritize proximal terrain features, thereby enhancing foothold safety.}
\label{fig8}
\end{figure}

\begin{figure}[t]
\centering  
\subfigure[]{
\label{fig9-1}
\includegraphics[width=4.2cm]{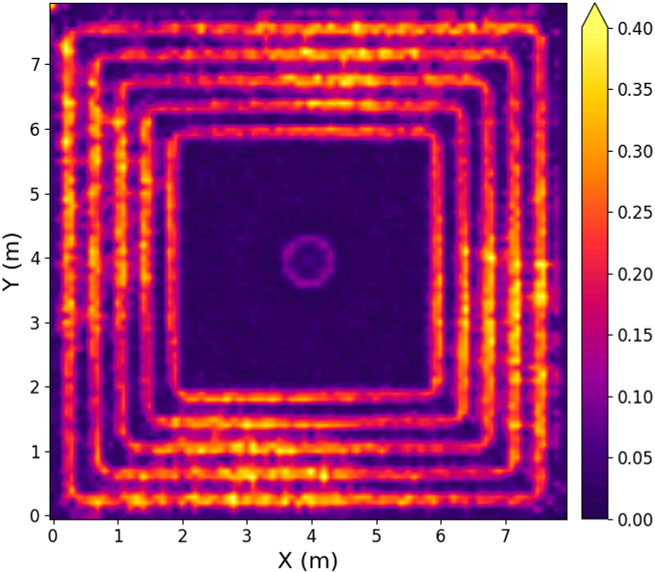}}
\subfigure[]{
\label{fig9-2}
\includegraphics[width=4.2cm]{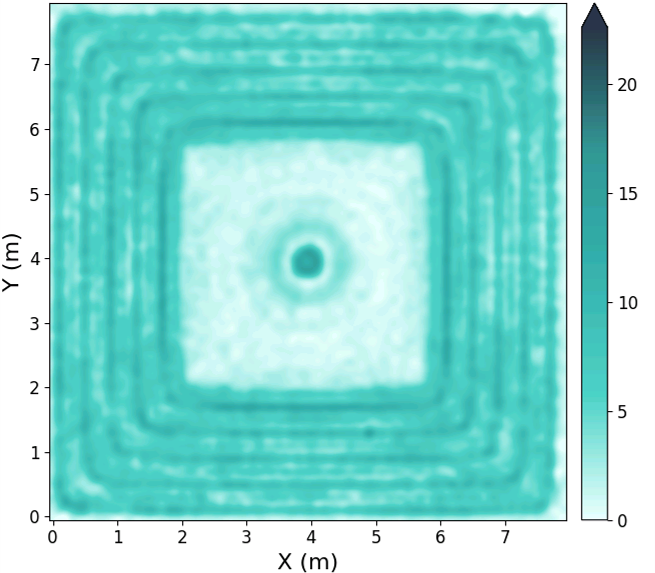}}
\caption{Correlation between SASG and motion stability.
(a) Spatial distribution of attention gating truncation rate.
The truncation rate increases significantly as the robot transitions from central flat ground to surrounding stairs.
(b) Stability improvement attributed to SASG.
Comparative analysis of spatial stability distributions demonstrates that SASG significantly enhances FocusNav’s performance on stair terrain relative to the WGSCA-Only baseline.}
\label{fig9}
\end{figure}

\subsubsection{Detailed Analysis of the SASG}\label{Experiments-2-3}
To further investigate the impact of the SASG module, we evaluate its dynamic response in challenging, unstable terrains.
As illustrated in Fig.~\ref{fig8}, we visualize the attention gating signals of the robot navigating through dynamic and complex terrains.
The spatio-temporal relationship between the robot’s position and the gating signals demonstrates that during stable locomotion on flat ground, the gating mechanism remains fully open, allowing the integration of long-range waypoint features for global navigation.
However, when the robot encounters unstructured terrain disturbances that compromise its stability, the SASG module rapidly truncates environmental information from distal waypoints.
This prevents distant obstacles from interfering with motion decisions, thereby mitigating potential stability issues during navigational maneuvers.
This dynamic selective attention compels the model to prioritize proximal spatial features in complex terrains, enhancing foothold safety.
Consequently, the mechanism achieves an optimal balance between short-term motion stability and global goal-directed perception.
Furthermore, we visualize the correlation between the SASG module and the robot's motion stability.

As illustrated in Fig.~\ref{fig9-1}, we present the spatial distribution of the attention gating truncation rate.
Specifically, by having the robot navigate from a central flat region across surrounding staircase areas, we recorded the frequency of gating activations at different spatial coordinates to calculate the spatial distribution across various terrains.
The results indicate that the gating activation rate remains low in the flat central region but increases significantly as the robot transitions to the staircases.
This trend aligns with the human-like heuristic that prioritizes local terrain details in complex areas to ensure foothold safety.
Fig.~\ref{fig9-2} demonstrates the percentage of stability improvement achieved by FocusNav compared to the WGSCA-Only baseline.
The positive correlation between stability improvement and SASG activation suggests that without SASG, the robot tends to execute aggressive navigational maneuvers influenced by distal obstacles, which often leads to falls on unstructured terrains.
In contrast, FocusNav prioritizes local stability when necessary, thereby significantly improving the overall completion rate of long-range navigation tasks.

\begin{figure*}[t]
\centerline{\includegraphics[width=18cm]{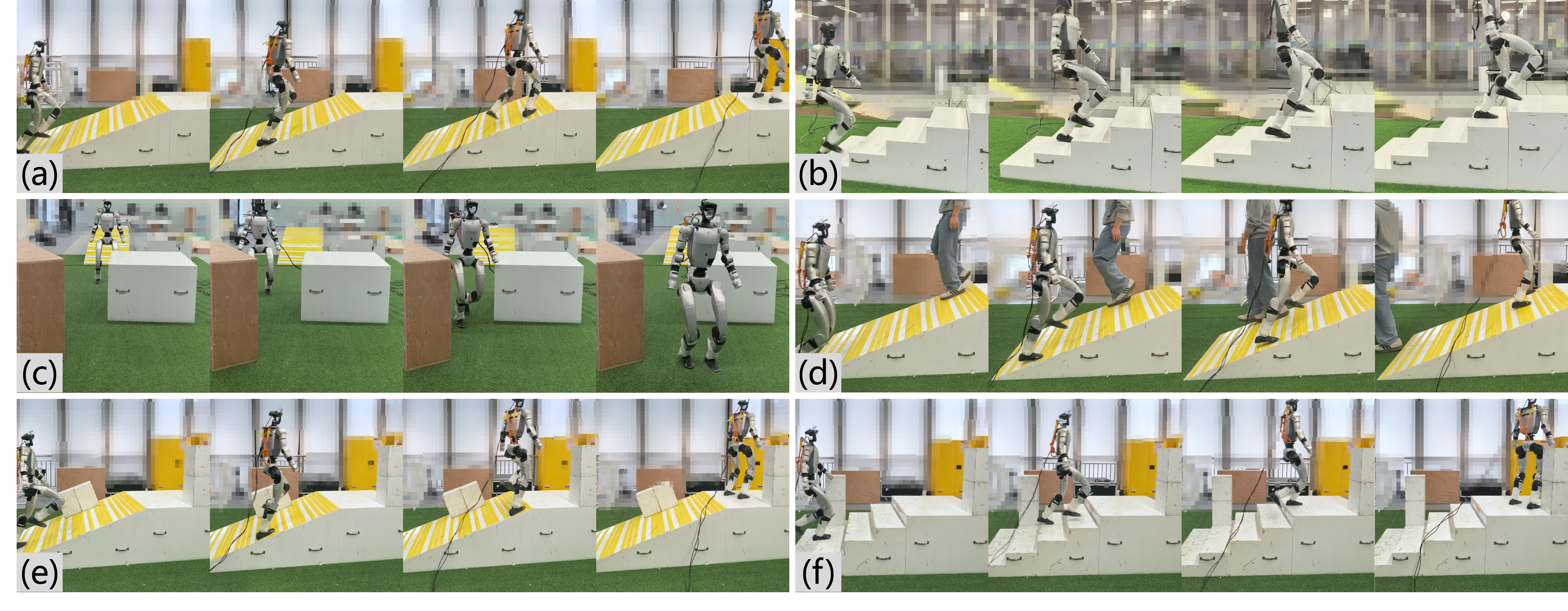}}
\caption{Robust local navigation of FocusNav in diverse and challenging environments.
(a) Traversing $22^\circ$ slopes.
(b) Traversing $16\,\text{cm}$ stairs.
(c) Local navigation through static obstacles on flat ground.
(d) Concurrent $22^\circ$ slope traversal and dynamic obstacle avoidance.
(e) Navigating through static obstacles on $22^\circ$ slopes.
(f) Navigation across $16\,\text{cm}$ stairs with static obstacles.
The results highlight the effectiveness of FocusNav in maintaining motion stability while executing navigational maneuvers in unstructured and dynamic settings.}
\label{fig10}
\end{figure*}

\begin{figure}[t]
\centerline{\includegraphics[width=8.7cm]{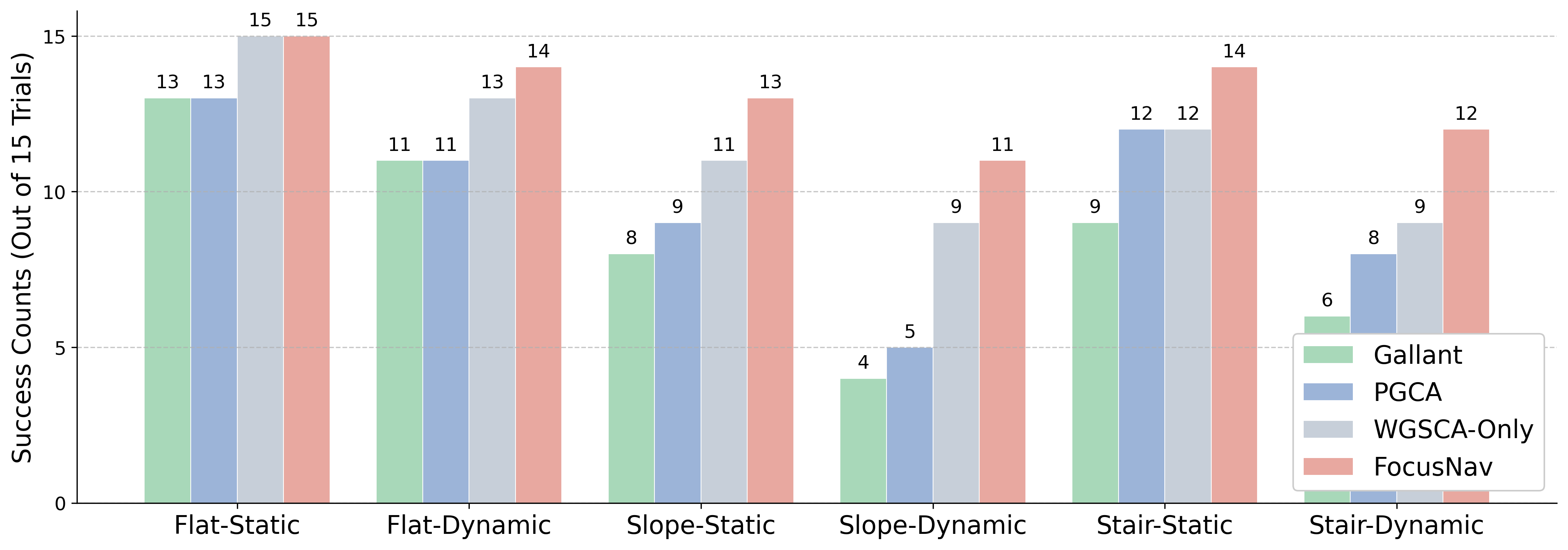}}
\caption{Navigation success rates across 15 trials in real-world dynamic and complex environments.
FocusNav outperforms all baselins.
Benefiting from the dual-layer spatial selective attention mechanism, FocusNav effectively handles both dynamic obstacles and complex terrains, substantially enhancing the navigational performance of the humanoid robot.}
\label{fig11}
\end{figure}

\subsection{Real-World Experiments}\label{Experiments-3}
FocusNav is deployed on a Unitree G1 humanoid robot to evaluate the practical performance of local navigation in unstructured and dynamic environments.
We establish a standardized testing arena that features a variety of challenging terrains and obstacles, including $16\,\text{cm}$ stairs, $22^\circ$ slopes, static box obstacles, and dynamic pedestrian interference.
Based on the combinations of terrain and obstacle types, the experimental configurations are categorized into six distinct scenarios.
Snapshots of FocusNav successfully traversing these diverse complex environments are illustrated in Fig.~\ref{fig10}.
We compare FocusNav against several baseline methods.
For each scenario, we conduct 15 independent trials to record the number of successful missions, defined as reaching the navigation target point without falling or colliding. 
As shown in Fig.~\ref{fig11}, FocusNav significantly outperforms all baseline methods across all test cases.
In highly demanding tasks, such as traversing narrow $16\text{cm}$ stairs amidst dynamic obstacles, FocusNav demonstrates the capability to plan collision-free paths for pedestrian avoidance while maintaining precise foothold placement, as illustrated in Fig.~\ref{fig1}.
FocusNav’s superior performance stems from its dual-layer spatial selective attention.
By anchoring perception to predicted navigational intent via WGSCA, the framework filters task-irrelevant noise to maintain focus on collision-free trajectories.
Concurrently, the SASG module dynamically prioritizes proximal terrain features over distal cues during instability to ensure foothold safety.
This hierarchical perception achieves an optimal balance between goal-directed navigation and motion stability, significantly enhancing performance in complex real-world settings.

\section{Conclusion}\label{Conclusion}
In this work, we propose FocusNav, a novel spatial selective attention framework designed to address the dual challenges of navigation and stability for humanoid robots in unstructured environments.
By integrating the WGSCA mechanism with the SASG module, our approach enables the robot to dynamically shift its perceptual focus between global navigation targets and local terrain features in response to real-time dynamic stability.
Experimental results on the Unitree G1 humanoid robot validate that the proposed method effectively enhances motion robustness and collision avoidance capabilities in complex navigation scenarios.
Our work advances the practical application of robust humanoid navigation, demonstrating that spatial selective perception is a key enabler for agile locomotion.
By bridging the gap between intentional guidance and reactive stability, FocusNav paves the way for the deployment of versatile humanoid robots in complex, real-world applications.

\section{Limitations}\label{Limitations}
Despite the superior performance of FocusNav, several limitations remain to be addressed in future research.
FocusNav primarily concentrates on the forward direction and lacks comprehensive focus on the robot's immediate surroundings.
This limits its applicability in scenarios requiring backward maneuvers or those with obstacles situated behind the robot.
Future work could focus on integrating proactive spatial region selection modules to enhance omnidirectional perceptual awareness.
FocusNav requires pre-defined navigational goals and lacks high-level semantic understanding for long-chain navigation tasks.
Future research will explore the integration of Vision-Language Models (VLMs) to enable socially-aware navigation and complex reasoning within human-centric environments.



\bibliographystyle{plainnat}


\end{document}